\documentclass{article}

\usepackage[utf8]{inputenc}
\usepackage[T1]{fontenc}
\usepackage{hyperref}
\usepackage{url}
\usepackage{booktabs}
\usepackage{amsfonts}
\usepackage{nicefrac}
\usepackage{microtype}
\usepackage{xcolor}
\usepackage{amsmath}
\usepackage{graphicx}
\usepackage{enumitem}
\usepackage{subcaption}
\usepackage{float}
\usepackage{algorithm}
\usepackage{algorithmic}
\usepackage{natbib}
\graphicspath{{figures/plots/}}

\title{
Brain-Language-Action (BLA) Models: Language-Conditioned EEG for Robotics Control
}

\author{
  Alexandr Plashchinsky \\
  VECTOR Labs \\
  San Francisco Bay Area \\
  \texttt{aplashch@gmail.com}
}

\begin{document}

\maketitle

\begin{abstract}
Electroencephalography (EEG)-based robotic control is commonly formulated as a direct classification problem, in which electrical neural signals are mapped to a fixed set of discrete actions. However, the limited separability and high noise of EEG signals make it difficult to scale this approach to fine-grained robotic control spaces. We introduce Brain-Language-Action (BLA) models, a framework in which language conditions the interpretation of neural representations for robotic action generation. In a BLA, a small set of reliably distinguishable brain states can be dynamically associated with different actions through a language-defined control mapping, allowing a small number of neural classes to apply to a larger global action space.

We develop a proof-of-concept BLA for drone control using motor-imagery EEG from the BCI Competition IV 2a dataset. The system is trained in two stages. First, we evaluate multiple candidate EEG encoder architectures using subject-specific four-class motor-imagery classification, converting 250Hz, 3.5-second, 22-channel EEG samples into five 128-dimensional brain-token embeddings. Second, these embeddings are projected into the embedding space of a pretrained large language model (LLM) and jointly fine-tuned with language instructions to autoregressively generate structured three-token drone actions. Across 840 possible language-defined mappings between four neural states and seven flight action combinations, the resulting BLA achieves 90\% per-token accuracy during evaluation. These results provide an initial demonstration that language conditioning can expand the effective control range of EEG-based robotic interfaces without requiring a corresponding increase in the number of directly distinguishable neural states.
\end{abstract}

\section{Introduction}

Brain-computer interfaces (BCIs) seek to translate neural activity into signals that can be used to communicate with or control external systems\citep{wolpaw2002bci}. Electroencephalography (EEG) is particularly attractive for this setting because it is non-invasive, relatively inexpensive, and capable of capturing neural activity at high temporal frequencies. Motor-imagery EEG, in which a user imagines performing a physical movement without executing it, has consequently become a widely studied paradigm for neural control. In a conventional motor-imagery BCI, a window of EEG data is processed by a machine learning model and classified into one of a fixed number of mental states, which can then be associated with commands for a downstream program\citep{khosla2020eeg,pfurtscheller2001motor}.

\paragraph{}
Deep neural networks have substantially improved the ability to learn these mappings directly from EEG signals. Convolutional neural networks (CNNs) have been used to extract temporal and spatial patterns from EEG windows, reducing the reliance on manually engineered features. It has been demonstrated that deep convolutional architectures can decode imagined and executed movements directly from EEG while outputting meaningful feature data\citep{schirrmeister2017deep}. EEGNet later introduced a compact architecture based on depthwise and separable convolutions that generalized across multiple BCI paradigms\citep{lawhern2018eegnet}. More recently, EEG-Inception applied multi-scale Inception-style convolutional processing to EEG data and demonstrated strong performance on the four-class BCI Competition IV 2a benchmark\citep{zhang2021eeginception}. Collectively, these methods demonstrate that useful discriminative representations can be learned directly from multichannel EEG data across time.

\paragraph{}
Despite these advances, EEG-based control remains constrained by the difficulty of separating a large number of neural classes. EEG measurements are noisy, subject-dependent, and susceptible to both physiological and external artifacts\citep{hameed2025motorimagery,padfield2019motorimagery}. Representative motor-imagery benchmarks therefore operate over relatively small fixed label spaces. For example, BCI Competition IV 2a contains four motor-imagery classes corresponding to the left hand, right hand, feet, and tongue\citep{tangermann2012bci}. When such a classifier is used directly for robotic control, the semantics of each predicted class are typically fixed: if left-hand imagery represents one action and right-hand imagery represents another, increasing the number of independently addressable robot commands requires either additional distinguishable neural states or an additional mechanism for specifying what those states should mean. This creates a practical bottleneck for systems intended to control robots with broader action spaces.

\paragraph{}
Modern robotic learning has approached the problem of large and diverse action spaces from a different direction. Vision-Language-Action (VLA) models combine perceptual representations with natural-language instructions and generate robot actions from a unified computational model. RT-2, for example, extends pretrained vision-language models to robotic control by representing robot actions as tokens and co-fine-tuning the model on vision-language and robotic action data. This formulation allows semantic knowledge encoded by large pretrained models to participate directly in action generation rather than treating language understanding and low-level control as entirely separate systems. The success of VLA architectures suggests that language can provide a powerful conditioning signal for interpreting other modalities in the context of robotic action\citep{zitkovich2023rt2}.

\paragraph{}
In this work, we investigate an analogous formulation for neural interfaces and introduce \textbf{Brain-Language-Action (BLA) models}. Rather than defining a permanent one-to-one mapping between an EEG class and a robot command, a BLA uses language to specify the current semantics of a set of learned neural states. Given an EEG representation $E$ and a language instruction $L$, the model generates an action sequence
\begin{equation}
    A = f(E, L),
\end{equation}
such that the interpretation of $E$ is conditioned on $L$. A neural state therefore does not need to correspond to the same robotic action in every context. For example, left-hand motor imagery may indicate forward motion under one language-defined control mapping and vertical ascent under another. The number of simultaneously distinguishable neural states remains unchanged, but those states can be reused across contexts to address a larger global space of robotic actions.

\paragraph{}
We demonstrate this idea with a proof-of-concept BLA for drone control. We first pretrain and compare multiple EEG encoder architectures on subject-specific four-class motor-imagery classification using BCI Competition IV 2a. Each encoder converts a 250Hz, 3.5-second, 22-channel EEG window into a sequence of five numeric embeddings. The strongest encoder is then integrated with a pretrained large language model (LLM) through a learned brain-to-language multilayer perceptron (MLP). The projected EEG embeddings are prepended to the language-model input, allowing the model to jointly process the neural signal and textual specification of the current control mapping. The BLA autoregressively generates three structured action tokens representing horizontal, rotational, and vertical drone control.

\paragraph{Contributions.}
The primary contributions of this work are:
\begin{itemize}[leftmargin=*,nosep]
    \item We introduce \textbf{Brain-Language-Action models}, a formulation for robotic control in which language conditions the interpretation of learned neural representations.
    \item We implement a proof-of-concept drone controller in which four motor-imagery EEG states can be dynamically assigned across seven validated robotic action sequences through 840 language-defined control mappings.
\end{itemize}

\section{Brain-Language-Action Models}

A \textbf{Brain-Language-Action (BLA) model} is a multimodal machine learning architecture in which neural activity (EEG) and language are jointly used to generate robotic actions. At a high level, a BLA consists of three components: a brain encoder, a brain-to-language projection module, and a pretrained language model. Given an EEG window $X \in \mathbb{R}^{T \times C}$, where $T$ denotes timesteps and $C$ denotes EEG channels, the brain encoder produces a set of neural embeddings
\begin{equation}
    E = B_{\theta}(X),
\end{equation}
which are subsequently projected into the embedding dimensionality of the internal language model,
\begin{equation}
    Z = P_{\phi}(E).
\end{equation}
The projected brain embeddings are then concatenated with the token embeddings of a language instruction $L$ and processed jointly by the language model. The resulting policy can be expressed as
\begin{equation}
    A = f(Z,L),
\end{equation}
where $A$ is a sequence of autoregressively generated robotic action tokens.

\paragraph{}
The central distinction between a BLA and conventional EEG-based robotic control is that the action mapping of a neural state is not permanently fixed. In a direct EEG-to-action classifier, a decoded neural class $c$ is typically associated with a fixed robotic action $a=g(c)$. In a BLA, the mapping becomes conditioned on language,
\begin{equation}
    a = g(c,L),
\end{equation}
allowing the same neural state to represent different actions under different linguistic contexts. This does not increase the number of simultaneously distinguishable neural states, but rather allows a small cluster of distinct neural states to be reused across a larger global action space. For example, assigning four distinguishable neural states to four distinct actions selected from a seven-action-sequence vocabulary yields
\begin{equation}
    P(7,4)=7\times6\times5\times4=840
\end{equation}
possible language-defined control mappings.

\paragraph{}
Our implementation follows an action-token formulation inspired by RT-2-style Vision-Language-Action models \citep{zitkovich2023rt2}. A fixed number of action tokens is generated at every control step, with each token position corresponding to one controllable degree of freedom of the robot. Each autoregressive token position draws from a shared set of discrete action values, allowing the generated token sequence to represent a structured multi-dimensional control command. The model predicts only the action for the next control step rather than an action trajectory, and does not require previous robot actions or the current robot state as input.

\paragraph{}
Training follows two stages. First, the brain encoder is pretrained using an EEG classification objective. A lightweight classifier is attached to the encoder output during this stage so that the quality of the learned neural representations can be evaluated through classification performance. After pretraining, this classifier is removed and only the encoder-generated embeddings are retained. The brain encoder, brain-to-language projection module, and pretrained language model are then co-fine-tuned for autoregressive robotic action generation.

\section{Brain Encoder Pretraining}

The first stage of our BLA implementation involves generating compact representations of the motor-imagery EEG that is independent of any particular robotic action space. Rather than training the brain encoder directly as part of the language-conditioned control objective, we first pretrain and evaluate candidate encoder architectures using EEG classification. The objective of this stage is to identify an encoder capable of transforming raw multichannel EEG windows into compact, discriminative neural embeddings that can subsequently be integrated into the BLA architecture.

\subsection{Dataset and Windowing}

We use the BCI Competition IV 2a motor-imagery EEG dataset \citep{tangermann2012bci}. The dataset contains recordings from nine subjects performing four motor-imagery tasks: left-hand movement, right-hand movement, feet movement, and tongue movement. EEG was recorded from 22 electrodes at a sampling frequency of 250 Hz.

\paragraph{}
Training is performed independently for each subject rather than across subjects. Each EEG sample consists of a non-overlapping 3.5-second window, corresponding to 875 timesteps across 22 EEG channels. The resulting encoder input is therefore represented as

\begin{equation}
    X \in \mathbb{R}^{875 \times 22}.
\end{equation}

Subject-specific training is used throughout this stage so that differences in EEG patterns between subjects are not conflated with differences between candidate encoder architectures. Consequently, one model is trained for every combination of subject and candidate encoder architecture.

\subsection{Candidate Brain Encoders}

We evaluate nine candidate encoder architectures spanning several common neural-network families: a flattened multilayer perceptron (MLP), recurrent neural network (RNN), long short-term memory network (LSTM), gated recurrent unit (GRU), convolutional neural network (CNN), subtractive CNN, Transformer encoder, subtractive Transformer encoder, and an Inception-based EEG encoder.

\paragraph{}
The Inception encoder is adapted from the motor-imagery EEG-Inception architecture implemented in Braindecode, which applies parallel multi-scale convolutional processing to EEG signals \citep{zhang2021eeginception,braindecode}. The subtractive CNN and subtractive Transformer variants use two instances of their respective feature-extraction structures and subtract their outputs through a learned operation intended to suppress noisy signal components.

\subsection{Common Embedding Representation}

To ensure that downstream classification performance reflects the quality of the learned EEG representation rather than differences in output dimensionality, every candidate encoder is constrained to produce the same representation, with the same patched input format.

\paragraph{}
Each 875-timestep EEG window is divided into five consecutive 175-timestep patches,

\begin{equation}
    X = \{X_1, X_2, X_3, X_4, X_5\},
\end{equation}

where each patch contains all 22 EEG channels. The candidate encoder independently transforms each patch into a 128-dimensional embedding,

\begin{equation}
    e_i = B_{\theta}(X_i), \qquad e_i \in \mathbb{R}^{128},
\end{equation}

producing the complete brain representation

\begin{equation}
    E = [e_1,e_2,e_3,e_4,e_5] \in \mathbb{R}^{5 \times 128}.
\end{equation}

This fixed representation interface is maintained across all candidate architectures. The five embeddings constitute the brain-token representation later supplied to the BLA model.

\subsection{Classification Pretraining}

During pretraining, a lightweight perceptron-based classifier is attached to the output of each brain encoder. The classifier maps the common $5 \times 128$ representation to four logits corresponding to the four motor-imagery classes,

\begin{equation}
    \hat{y} = C_{\psi}(E), \qquad \hat{y} \in \mathbb{R}^{4}.
\end{equation}

The classifier architecture is identical for every candidate encoder. This intentionally limits the analytical contributions of the task-specific prediction head and provides a common evaluation mechanism across architectures. Under this formulation, successful classification requires the encoder itself to produce representations containing sufficient discriminative information for a comparatively simple classifier to determine the corresponding motor-imagery class.

\paragraph{}
Each subject-specific model is trained for 30 epochs with a learning rate of $10^{-3}$ and batch size of 16. For each subject, 10\% of the available windowed samples are held out for evaluation. Random initialization and dataset splitting use a fixed seed of 2025.

\paragraph{}
Following pretraining, the classifier head is discarded and only the EEG encoder is retained. Across the candidate architectures, our adapted EEG-Inception encoder produced the strongest classification performance and was therefore selected as the brain encoder for the downstream BLA experiments. Quantitative comparisons between encoder architectures and subject-specific performance are reported in the following sections.

\section{BLA Architecture and Fine-Tuning}

Following brain-encoder pretraining, we construct the complete BLA as a language-conditioned drone control model. The selected EEG-Inception encoder produces five 128-dimensional brain embeddings from each EEG window. These embeddings are integrated with a pretrained Qwen3-0.6B language model \citep{qwen3technicalreport} through a learned brain-to-language projection module. Four subject-specific BLA models are trained using subjects 1, 3, 7, and 8, for which the selected brain-encoder architecture produced its strongest classification performance.

\subsection{Brain-to-Language Projection}

For an EEG input $X$, the pretrained brain encoder produces

\begin{equation}
    E = B_{\theta}(X), \qquad
    E \in \mathbb{R}^{5 \times 128}.
\end{equation}

Qwen3-0.6B uses a token-embedding dimensionality of 1024. We therefore introduce a multilayer perceptron (MLP) projection module $P_{\phi}$ that transforms each 128-dimensional brain embedding into the language-model embedding space,

\begin{equation}
    Z = P_{\phi}(E), \qquad
    Z \in \mathbb{R}^{5 \times 1024}.
\end{equation}

Unlike the brain encoder, the projection module is not independently pretrained. It is initialized as part of the complete BLA and learned during BLA fine-tuning.

\paragraph{}
For a language instruction $L$, the projected brain embeddings are prepended to the token embeddings of the instruction. The resulting multimodal sequence is therefore

\begin{equation}
    H = [Z;\operatorname{Embed}(L)],
\end{equation}

which is provided directly to the language model. The brain encoder, projection module, and language model are then co-fine-tuned for autoregressive action generation.

\subsection{Drone Action Representation}

The BLA produces a fixed sequence of three action tokens representing three independently controllable dimensions of drone motion:

\begin{equation}
    A =
    [a_{\mathrm{horizontal}},
     a_{\mathrm{yaw}},
     a_{\mathrm{vertical}}].
\end{equation}

The first token controls forward and backward movement, the second controls left and right yaw (turning left and right), and the third controls ascent and descent. Each token position draws from the same set of three discrete action tokens,

\begin{equation}
    a_i \in
    \{\texttt{ACT\_0},\texttt{ACT\_1},\texttt{ACT\_2}\},
\end{equation}

where \texttt{ACT\_0} represents negative motion along the corresponding control dimension, \texttt{ACT\_1} represents a neutral command, and \texttt{ACT\_2} represents positive motion. Under this convention, backward movement, right yaw, and descent are negative actions, while forward movement, left yaw, and ascent are positive actions. The three action tokens are added to the language-model tokenizer as additional discrete tokens.

\paragraph{}
The seven action sequences used in this proof-of-concept are shown in Table~\ref{tab:drone_actions}.

\begin{table}[t]
\centering
\caption{Structured action-token representation used for drone control.}
\label{tab:drone_actions}
\begin{tabular}{lccc}
\toprule
\textbf{Action} & \textbf{Horizontal} & \textbf{Yaw} & \textbf{Vertical} \\
\midrule
Forward    & ACT\_2 & ACT\_1 & ACT\_1 \\
Backward   & ACT\_0 & ACT\_1 & ACT\_1 \\
Turn left  & ACT\_1 & ACT\_2 & ACT\_1 \\
Turn right & ACT\_1 & ACT\_0 & ACT\_1 \\
Ascend     & ACT\_1 & ACT\_1 & ACT\_2 \\
Descend    & ACT\_1 & ACT\_1 & ACT\_0 \\
Hover      & ACT\_1 & ACT\_1 & ACT\_1 \\
\bottomrule
\end{tabular}
\end{table}

\paragraph{}
The output represents a single subsequent control step rather than a multi-step action trajectory. The three tokens are generated autoregressively, such that each action token is predicted sequentially by the language model.

\subsection{Language-Conditioned Control Mappings}

Each training example combines an EEG window with a natural-language instruction specifying how the four motor-imagery states should be interpreted as drone actions. The instruction defines mappings for left-hand, right-hand, feet, and tongue imagery. For example, the text instructions may specify that left-hand imagery corresponds to forward motion while right-hand imagery corresponds to ascent. Under a different instruction, the same neural states may be assigned different robotic actions.

\paragraph{}
Formally, let

\begin{equation}
    \mathcal{C} =
    \{c_{\mathrm{left}},c_{\mathrm{right}},
      c_{\mathrm{feet}},c_{\mathrm{tongue}}\}
\end{equation}

denote the four EEG classes and let $\mathcal{A}$ denote the seven drone action sequences. 

\paragraph{}
Each language instruction defines a translative mapping

\begin{equation}
    m_L : \mathcal{C} \rightarrow \mathcal{A},
\end{equation}

where $L$ denotes the textual representation of the mapping. Because four distinct neural classes are assigned to four distinct actions selected from a vocabulary of seven, the complete set contains

\begin{equation}
    P(7,4)=840
\end{equation}

possible language-defined control mappings.

\paragraph{}
For an EEG window belonging to neural class $c$ and an instruction defining mapping $m_L$, the target action sequence is

\begin{equation}
    A=m_L(c).
\end{equation}

The training objective therefore requires the BLA to use both modalities jointly: the EEG input provides evidence for the current neural state, while the language instruction determines the robotic meaning assigned to that state. Neither input alone is sufficient to determine the correct action across the complete set of mappings.

\subsection{Subject-Specific Fine-Tuning}

BLA fine-tuning uses EEG data from the same BCI Competition IV 2a subjects used during brain-encoder pretraining. Based on the encoder evaluation, subjects 1, 3, 7, and 8 are selected for downstream BLA experiments, producing four independently trained BLA models.

\paragraph{}
Training remains strictly subject-specific. For subject $s$, the BLA is initialized using the brain encoder pretrained exclusively on subject $s$ and is subsequently fine-tuned using EEG samples from the same subject. EEG data from other subjects are not introduced into that model's training or evaluation pipeline.

\paragraph{}
Each BLA is fine-tuned for 30 epochs using a batch size of 16 and learning rate of $10^{-3}$. A 10\% subject-specific split is held out for evaluation, and a fixed random seed of 2025 is used for dataset splitting and random initialization. During fine-tuning, the pretrained brain encoder, brain-to-language projection MLP, and Qwen3-0.6B language model are optimized jointly to autoregressively generate the corresponding three-token drone action sequence.

\subsection{Action Sequence Simulation}

To validate the robotic action space used for BLA fine-tuning, we simulated the complete set of language-defined control mappings using the open-source Webots robotics simulator \citep{michel2004webots}. Simulations were performed using the provided Mavic 2 Pro drone template. Across all 840 neural-state-to-action mappings, the corresponding structured action sequences were executed in simulation to verify that each mapping produced a valid drone control command. This simulation stage was used for action-space validation only and was not incorporated into BLA training or evaluation.

\section{Results}
\label{sec:results}

We report results in three stages. First, we compare candidate brain-encoder architectures using average subject-specific classification accuracy. Second, we isolate the strongest subject-specific variants of the best overall encoder architecture. Third, we evaluate the full BLA fine-tuning results for the corresponding subject-specific BLA models.

\subsection{Brain Encoder Pretraining Results}

\begin{figure}[t]
\centering
\includegraphics[width=\linewidth]{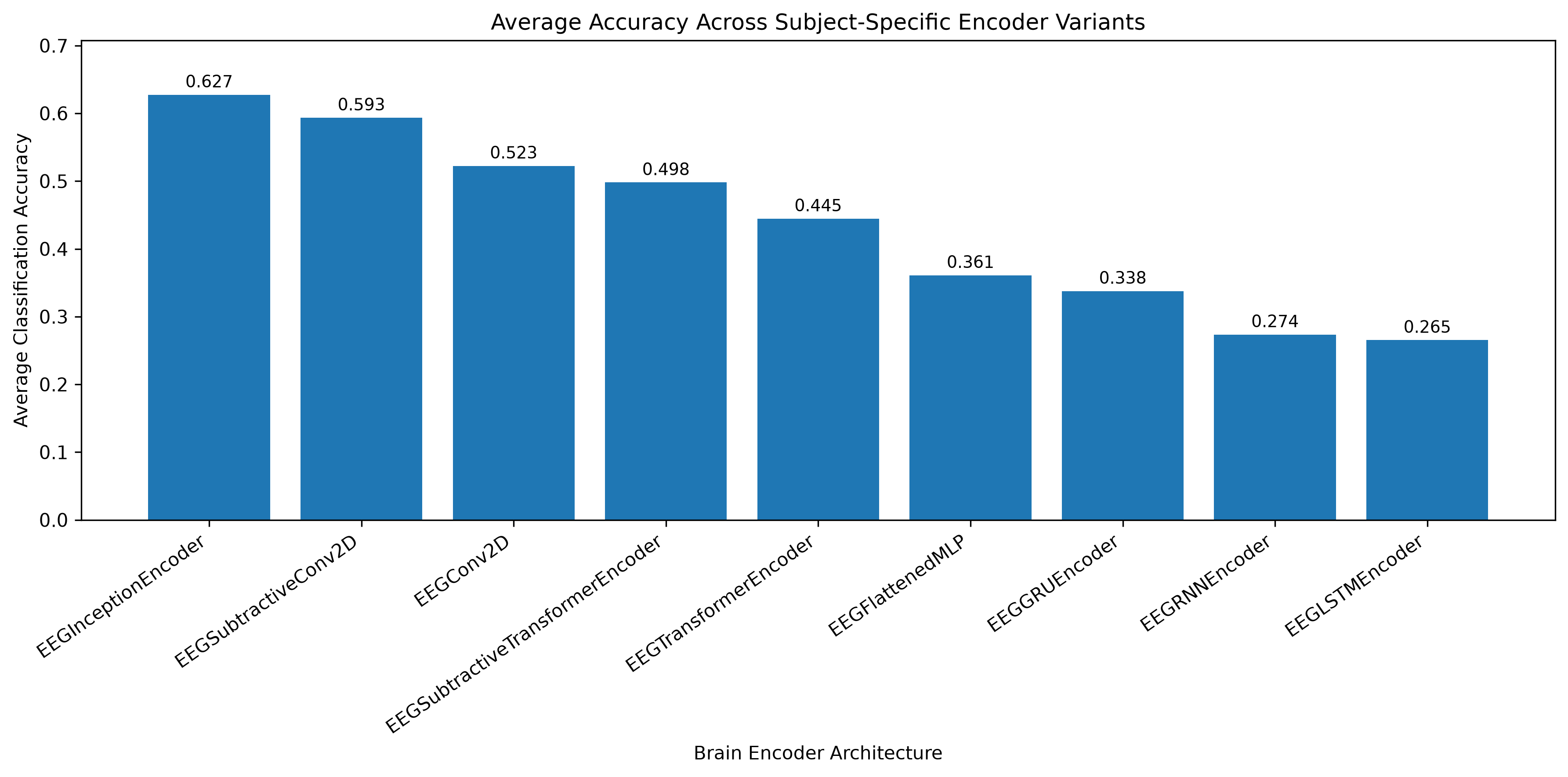}
\caption{Average motor-imagery classification accuracy for each candidate brain encoder, averaged across subject-specific model variants. The EEGInceptionEncoder achieved the strongest overall performance.}
\label{fig:encoder_avg_accuracy}
\end{figure}

Figure~\ref{fig:encoder_avg_accuracy} reports the average classification accuracy of each candidate brain encoder, averaged across all subject-specific model variants. Among all evaluated architectures, the EEGInceptionEncoder achieved the strongest overall performance with an average accuracy of 0.627. The next strongest architecture was the EEGSubtractiveConv2D (0.593), followed by EEGConv2D (0.523). Recurrent and standard Transformer-based architectures performed less strongly overall, with the lowest average accuracies observed for the EEGRNNEncoder (0.274) and EEGLSTMEncoder (0.265).

\paragraph{}
These results indicate that Inception-style convolutional processing was the most effective design for learning compact and discriminative EEG representations under the shared $5 \times 128$ embedding format. Accordingly, the EEGInceptionEncoder was selected as the brain encoder for downstream BLA fine-tuning.

\subsection{Best Subject Variants of the Selected Brain Encoder}

Although average performance across subjects provides the most reliable basis for architecture selection, EEG quality and class separability remain highly subject-dependent. To evaluate BLA performance under the strongest subject-specific EEG conditions, we retained the best-performing subject variants of the selected EEGInceptionEncoder.

\begin{figure}[t]
\centering
\includegraphics[width=\linewidth]{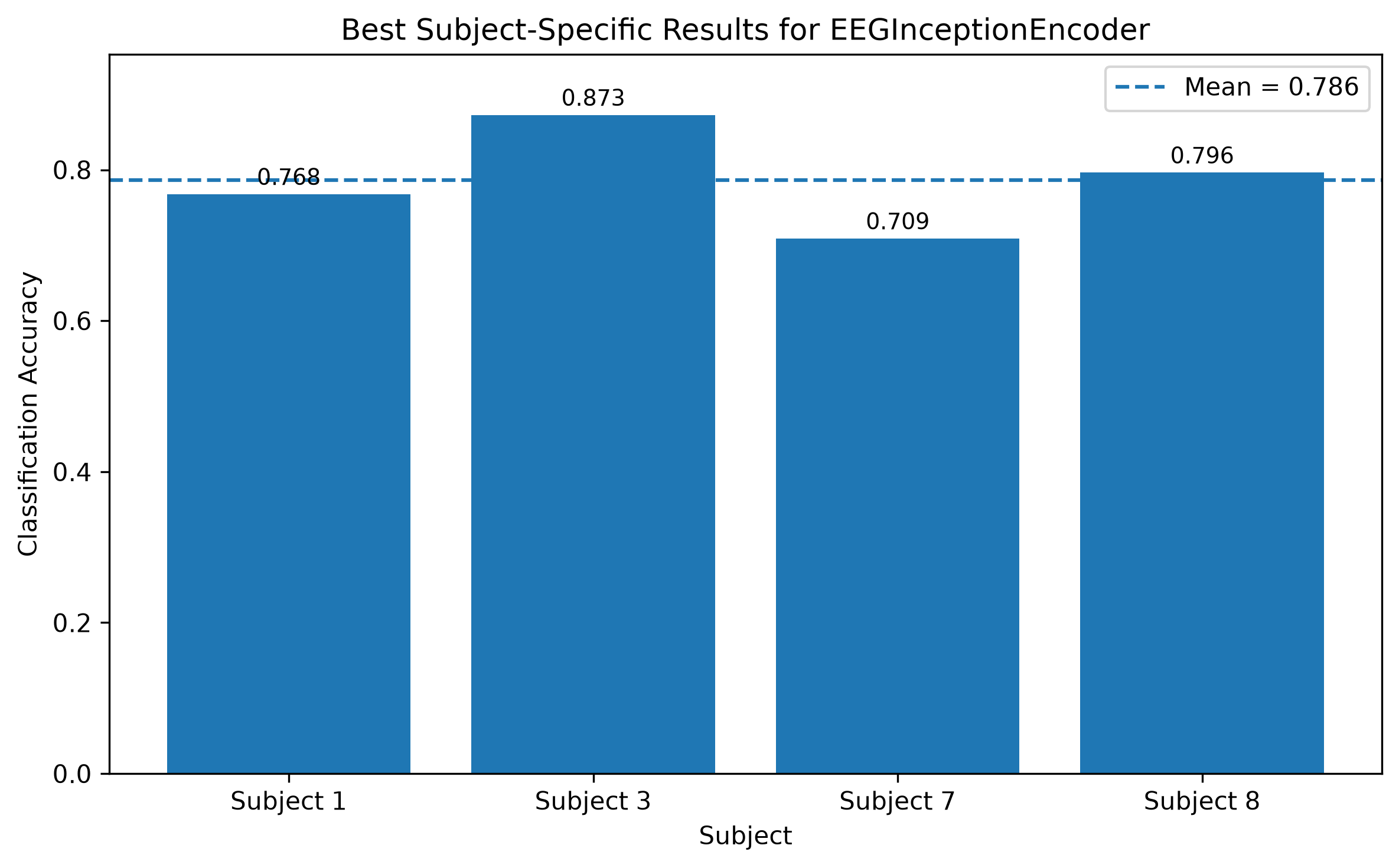}
\caption{Best subject-specific EEGInceptionEncoder results. Subjects 1, 3, 7, and 8 were selected for downstream BLA fine-tuning.}
\label{fig:inception_best_subjects}
\end{figure}

\paragraph{}
Figure~\ref{fig:inception_best_subjects} shows the four strongest subject-specific results for the EEGInceptionEncoder. The selected subjects were 1, 3, 7, and 8, with classification accuracies of 0.768, 0.873, 0.709, and 0.796, respectively. Across these four subject-specific models, the mean classification accuracy was 0.786. These checkpoints were then used to initialize the four brain encoders in the downstream BLA models.

\subsection{Brain-Language-Action Fine-Tuning Results}

\begin{figure}[H]
\centering
\includegraphics[width=\linewidth]{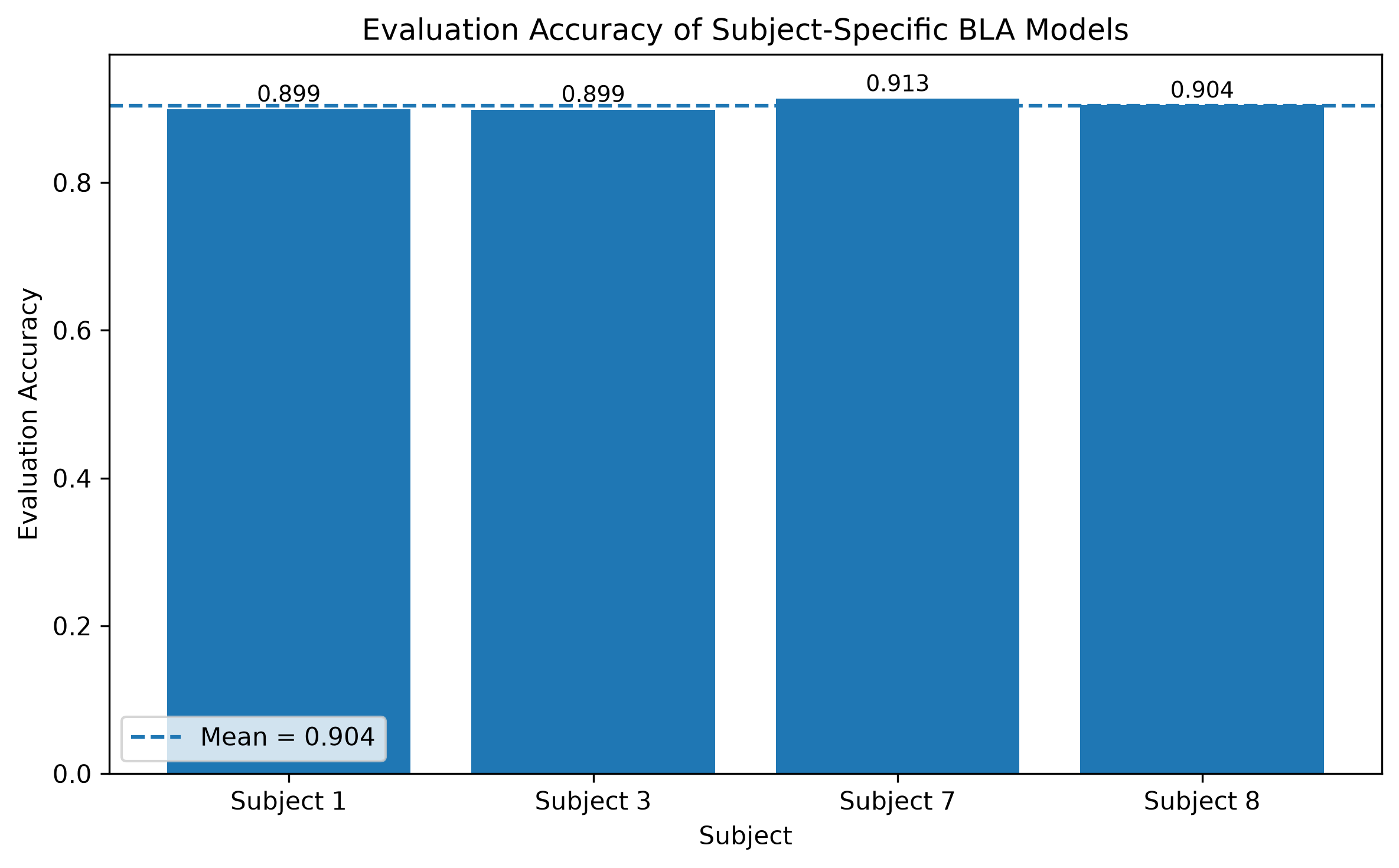}
\caption{Per-token evaluation accuracy of the four subject-specific BLA models. The average per-token BLA evaluation accuracy across subjects was 0.904.}
\label{fig:bla_subject_results}
\end{figure}

Figure~\ref{fig:bla_subject_results} reports the per-token evaluation accuracy of the four subject-specific BLA models. Each BLA was initialized using the pretrained EEGInceptionEncoder corresponding to the same subject and then jointly fine-tuned with the brain-to-language projection module and Qwen3-0.6B language model.

\paragraph{}
Across subjects 1, 3, 7, and 8, the resulting BLA models achieved accuracies of 0.899, 0.899, 0.913, and 0.905, respectively, corresponding to an average per-token evaluation accuracy of 0.904. Performance was therefore both high and consistent across all four subject-specific BLA variants.

\paragraph{}
These results provide initial empirical support for the BLA architecture. A compact EEG representation pretrained through motor-imagery classification can be successfully integrated with a language model and used to generate structured robotic action tokens with high accuracy under language-conditioned control mappings. This suggests that language can meaningfully expand the effective control range of EEG-based robotic interfaces without requiring a corresponding increase in the number of directly distinguishable neural states.

\section{Discussion and Limitations}

The results provide initial evidence that the Brain-Language-Action architecture is a viable approach for language-conditioned neural robotic control. Across the evaluated subject-specific models, the BLA was able to combine learned EEG representations with language-defined control mappings and accurately generate structured robotic action tokens. This supports the central hypothesis of this work: a limited set of distinguishable neural states can be reused across a broader robotic action space when language dynamically determines their interpretation. However, these experiments should be viewed as a proof of concept rather than a demonstration of general-purpose neural robotic control.

\paragraph{}
Many limitations remain. All experiments used a single motor-imagery EEG dataset and a single random seed, and the complete BLA architecture was evaluated using only one selected brain encoder architecture and one pretrained language model. The robotic experiments were restricted to drone control with three autoregressively generated action tokens and three discrete action bins per token. The architecture was also intentionally closer to the discrete action-token formulation of RT-2 than to more recent VLA designs such as $\pi_0$, which use continuous flow-matching policies conditioned on robot observations to generate action chunks \citep{black2024pi0}. Most importantly, all EEG models in this study were trained and evaluated independently for individual subjects. The resulting system therefore does not demonstrate cross-subject generalization or a foundational model capable of operating across users. Future work should evaluate BLA architectures across additional EEG datasets, subjects, robotic embodiments, language models, brain encoders, and more expressive continuous-action generation architectures.

\section{Conclusion}

We introduce Brain-Language-Action (BLA) models, a paradigm for robotic control in which language conditions the interpretation of learned EEG representations. Using a subject-specific proof-of-concept for drone control, we showed that compact brain embeddings can be projected into a pretrained language-model space and jointly used with textual control mappings to generate structured robotic action tokens with high evaluation accuracy. These results demonstrate the feasibility of using language to expand the effective control range of EEG-based encoders without requiring a larger number of directly distinguishable neural states. We are left with substantial room for future work in cross-subject generalization, broader robotic action spaces, and more foundational BLA architectures.

\clearpage
\bibliographystyle{plainnat}
\bibliography{references}

\clearpage
\appendix

\section{Experimental Configuration}

Table~\ref{tab:experimental_config} summarizes the primary experimental configuration used for brain-encoder pretraining and BLA fine-tuning.

\begin{table}[H]
\centering
\caption{Experimental configuration for the two training stages.}
\label{tab:experimental_config}
\begin{tabular}{lll}
\toprule
\textbf{Setting} & \textbf{Encoder Pretraining} & \textbf{BLA Fine-Tuning} \\
\midrule
Dataset & BCI Competition IV 2a & BCI Competition IV 2a \\
Subjects & 1--9 & 1, 3, 7, 8 \\
EEG Input & $875 \times 22$ & $875 \times 22$ \\
Brain Encoder & Candidate architectures & EEG-Inception \\
Brain Embeddings & $5 \times 128$ & $5 \times 128$ \\
Projection & -- & $128 \rightarrow 1024$ MLP \\
Language Model & -- & Qwen3-0.6B \\
Output & 4-class logits & 3 action tokens \\
Epochs & 30 & 30 \\
Batch Size & 16 & 16 \\
Learning Rate & $10^{-3}$ & $10^{-3}$ \\
Evaluation Split & 10\% & 10\% \\
Random Seed & 2025 & 2025 \\
\bottomrule
\end{tabular}
\end{table}

\paragraph{Model selection.}
Brain-encoder architecture selection was performed by comparing subject-specific motor-imagery classification performance across the candidate encoder families described in Section~\ref{sec:results}. The EEG-Inception architecture produced the strongest average performance across subjects and was therefore selected for the complete BLA architecture. Qwen3-0.6B was selected as the pretrained language model, with its 1024-dimensional token embedding space determining the output dimensionality of the brain-to-language projection MLP.

\paragraph{Compute.}
Brain-encoder pretraining was performed locally on an Apple MacBook Pro with an M4 Pro processor using the Metal Performance Shaders (MPS) driver. BLA fine-tuning was performed on two AWS \texttt{p4d.24xlarge} nodes, each containing eight NVIDIA A100 GPUs, for a total of 16 GPUs. Training was distributed across the 16 GPU workers using Ray Train, with gradients synchronized between workers through all-reduce during distributed optimization.

\paragraph{Code availability.}
The implementation used for brain-encoder pretraining, BLA fine-tuning, action-space generation, and evaluation is publicly available at:

\begin{center}
\url{https://github.com/alexplash/cerebrus-research-public.git}
\end{center}

\end{document}